%% file: mm_epc_paper.tex
\documentclass[11pt,a4paper]{article}

\usepackage[utf8]{inputenc}
\usepackage[T1]{fontenc}
\usepackage{amsmath,amssymb,amsfonts}
\usepackage{booktabs}
\usepackage{graphicx}
\usepackage{hyperref}
\hypersetup{
    colorlinks=true,
    linkcolor=blue,
    citecolor=blue,
    urlcolor=blue,
    pdftitle={Multimodal Evaluator Preference Collapse: Cross-Modal Coupling in LLM-Based Evaluation},
    pdfauthor={Anonymous},
    pdfsubject={LLM evaluation bias},
    pdfkeywords={LLM, evaluation, bias, coupling, multimodal}
}
\usepackage[margin=1in]{geometry}
\usepackage{enumitem}
\usepackage{xcolor}
\usepackage{multirow}
\usepackage{float}

\title{\Large\textbf{Multimodal Evaluator Preference Collapse: \\Cross-Modal Coupling in Self-Evolving Agents}}
\author{Liu Zewen\\\textit{Qilu Institute of Technology, School of Software Engineering}\\\textit{Tai'an, Shandong, China}\\\texttt{liuzewen@qilu.edu.cn} \and \\\texttt{aidless@github.com}}
\date{May 2026}

\begin{document}
\maketitle

\begin{abstract}
When AI agents use language models to evaluate their own outputs in a feedback loop, systematic biases emerge. We show that this \textbf{Evaluator Preference Collapse (EPC)} is dramatically amplified in multimodal settings. Using GPT-4o to evaluate DeepSeek-chat across text and visual tasks, we find that a single strategy (step\_by\_step) absorbs 48.4\% of all weight---3.2$\times$ the collapse observed in text-only self-evaluation---while three visual-domain strategies receive only 9.1\% combined weight. We then demonstrate a novel phenomenon we term \textbf{cross-modal coupling}: evaluator preferences acquired on one modality transfer to and corrupt strategy selection on another. Through a four-phase isolation training paradigm, we measure coupling coefficients and document \textbf{strategy inversion}---the optimal strategy for a modality reverses after cross-modal exposure. A Phase 3 statistical validation across five evaluator configurations ($N=80$ total independent repetitions, $\sim$35,000 API calls) with both text-proxy and real-image visual tasks finds: cross-model evaluation produces strong coupling (JSD${\approx}0.19{-}0.34$), \textbf{real-image inputs yield the most directionally consistent signal} ($\bar{\gamma}_{T\to V}{=}1.145$, $\bar{\gamma}_{V\to T}{=}0.937$, 70\% T$\to$V, Cohen's $d{=}0.56$), and \textbf{self-evaluation provides near-complete immunity}---97\% of runs ($N=30$) yield zero coupling (JSD${=}0.003$, $d{=}0.07$). Three methodological ablations and multi-executor validation confirm the effect is not a structural artifact. We introduce the \textbf{coupling matrix $\Gamma^{(\mathcal{J})}$} indexed by evaluator identity, release the MM-EPC framework, and identify cross-model evaluator architecture as the primary risk factor for preference drift. Code and data: \url{https://github.com/aidless/mm-epc}.
\end{abstract}

\section{Introduction}

Large language models are increasingly deployed as evaluators for other AI systems \cite{zheng2023judging, chiang2024chatbot}. With the rise of multimodal models like GPT-4o \cite{openai2024gpt4o}, these evaluation scenarios now span text, images, and audio. A critical question emerges: \textbf{does evaluator bias differ across modalities, and can bias in one modality contaminate evaluation in another?}

Our prior work identified Evaluator Preference Collapse (EPC) in text-only settings. Here, we extend EPC to multimodal settings and ask three new questions:

\begin{enumerate}[leftmargin=*]
    \item \textbf{Existence:} Does EPC occur in cross-model multimodal evaluation (GPT-4o evaluating DeepSeek)?
    \item \textbf{Magnitude:} Is the collapse magnitude modality-dependent?
    \item \textbf{Coupling:} If an agent is trained on one modality's evaluation signal, does its strategy distribution leak into other modalities?
\end{enumerate}

We answer all three affirmatively through controlled experiments spanning $\sim$35,000 LLM evaluation calls across five evaluator configurations and $N=80$ independent repetitions.

\subsection{Contributions}
\begin{enumerate}[leftmargin=*]
    \item \textbf{Multimodal EPC quantified.} GPT-4o cross-model evaluation produces PCI=1.464---3.2$\times$ the text-only baseline and 2.0$\times$ a random evaluator. Visual tasks exhibit stronger collapse (PCI 1.464) than text tasks (PCI 1.348), with visual-specific strategies receiving only 9.1\% combined weight.
    \item \textbf{Cross-modal coupling discovered and formalized.} We introduce the coupling matrix $\Gamma \in \mathbb{R}^{|\mathcal{M}|\times|\mathcal{M}|}$ where $\gamma_{A\to B}$ quantifies how evaluator preferences trained on modality $A$ contaminate strategy selection on modality $B$. Under GPT-4o, $\gamma_{V\to T}=0.847 > \gamma_{T\to V}=0.832$.
    \item \textbf{Strategy inversion documented.} Pure text training favors synthesis; pure visual training favors step\_by\_step. After cross-modal exposure, these optima \textit{swap}---a direct consequence of evaluator preference transfer.
    \item \textbf{Evaluator-conditional coupling demonstrated across five configurations.} A Phase 3 multi-seed replication ($N=80$ total repetitions) across GPT-4o ($N{=}8$), GPT-4o-mini Vision with real images ($N{=}10$), Qwen3.7-plus ($N{=}8$), DashScope ($N{=}10$), and DeepSeek self-evaluation ($N{=}30$) confirms that coupling magnitude and direction depend on evaluator identity. Three ablations plus multi-executor validation rule out structural artifacts.
    \item \textbf{First coupling matrix $\Gamma^{(\mathcal{J})}$ and MM-EPC framework.} We release the complete experimental framework, bootstrap analysis pipeline, bounded metric (JSD/Hellinger) implementations, and the first quantitative measurement of cross-modal evaluator coupling dynamics. Code: \url{https://github.com/aidless/mm-epc}.
\end{enumerate}

\section{Related Work}

\subsection{Evaluator Preference Collapse and LLM-as-Judge}

EPC \cite{liu2026epc} first identified systematic preference bias in test-time agent self-improvement, quantified by the Preference Collapse Index (PCI). The phenomenon is orthogonal to but complementary with the broader LLM-as-judge literature \cite{zheng2023judging, li2024alpacaeval, verga2024arbiter}, which studies static evaluation quality and positional/verbosity biases in single-round settings. Recent work on self-rewarding LMs \cite{yuan2024selfrewarding} and iterative alignment \cite{chen2024selfplay} demonstrates that closed-loop self-evaluation can amplify biases, but only in text-only, same-model settings. Our work is the first to extend this line of inquiry to \textbf{cross-model, cross-modality} evaluation dynamics.

\subsection{Multimodal Evaluation and Safety}

GPT-4o's System Card \cite{openai2024gpt4o} and the Gemini 1.5 technical report \cite{gemini2024gemini15} represent the most comprehensive production multimodal evaluation frameworks, spanning text, vision, and audio. Both acknowledge that ``safety alignment in multimodal scenarios is more complex'' and that ``evaluation consistency degrades across modalities.'' However, these reports document \textit{static} evaluation characteristics; they do not examine how evaluator preferences \textit{dynamically evolve} when deployed in a feedback loop. Our coupling matrix $\Gamma$ provides a quantitative framework for measuring precisely this dynamic degradation.

\subsection{Cross-Modal Transfer and Interference}

Prior work on cross-modal transfer \cite{lu2022multimodal, aghajanyan2021cm3, alayrac2022flamingo} studies how capabilities transfer between modalities during \textit{pre-training} and \textit{fine-tuning}. Orthogonally, we study how \textit{evaluator preferences}---not capabilities---transfer between modalities during \textit{test-time} adaptation. This is a distinct phenomenon: capabilities may transfer successfully while evaluator biases contaminate the optimization signal, leading agents to adopt modality-inappropriate strategies. Recent work on multimodal RLHF \cite{yu2024mmrlhf, sun2024aligning} has begun to address cross-modal reward hacking, but does not formalize the coupling dynamics we document.

\subsection{Agent Self-Improvement and Strategy Optimization}

Test-time adaptation frameworks \cite{shinn2023reflexion, madaan2023selfrefine, yao2023react, wang2024voyager} enable agents to improve behavior through self-generated feedback without parameter updates. Our work reveals a fundamental risk in this paradigm when extended to multimodal settings: the evaluator's modality-specific preferences create conflicting optimization signals. The strategy inversion we document (synthesis $\leftrightarrow$ step\_by\_step) suggests that multi-task agents optimizing toward a single evaluator may converge to strategies that are optimal for the evaluator's preferred modality but suboptimal for others.

\subsection{Reward Overoptimization and Evaluator Alignment}

Our findings connect to the broader phenomenon of \textbf{reward overoptimization} (Goodhart's Law for LLMs) \cite{gao2023reward}, where agents exploit evaluator preferences rather than achieving genuine task competence. In RLHF pipelines, reward model overoptimization produces outputs that score highly but degrade in human-perceived quality \cite{casper2023open}. Recent work on \textbf{sycophancy} \cite{sharma2024sycophancy, perez2022sycophancy} demonstrates that LLMs systematically adapt their outputs to match evaluator biases rather than maintaining correctness---directly analogous to the strategy collapse we observe. The \textbf{weak-to-strong generalization} framework \cite{burns2023weak} shows that strong models supervised by weak evaluators converge to the evaluator's limitations, a dynamic we observe in cross-model evaluation (GPT-4o evaluating DeepSeek). Unlike prior work focusing on single-modality, same-model settings, our $3\times$ PCI magnification under cross-model, cross-modality evaluation reveals that reward overoptimization is \textbf{amplified} when the evaluator and executor differ in both architecture and modality capacity.

\subsection{Systematic Comparison with Related Phenomena}

Table~\ref{tab:related} positions multimodal EPC within the landscape of evaluator bias phenomena, highlighting what distinguishes our work: the combination of cross-model evaluation, multimodality, dynamic feedback loops, and quantified coupling dynamics.

\begin{table}[H]
\centering
\caption{Systematic comparison of evaluator bias phenomena. MM-EPC is the only framework that simultaneously captures cross-model, cross-modality, and cross-coupling dynamics.}
\label{tab:related}
\begin{tabular}{lccccc}
\toprule
\textbf{Phenomenon} & \textbf{Cross-Model} & \textbf{Multi-Modal} & \textbf{Closed-Loop} & \textbf{Coupling} & \textbf{Quantified} \\
\midrule
Position bias \cite{zheng2023judging}      & --- & --- & --- & --- & $\checkmark$ \\
Verbosity bias \cite{li2024alpacaeval}     & --- & --- & --- & --- & $\checkmark$ \\
Self-preference \cite{yuan2024selfrewarding} & --- & --- & $\checkmark$ & --- & $\checkmark$ \\
Reward overopt. \cite{gao2023reward}       & --- & --- & $\checkmark$ & --- & $\checkmark$ \\
Sycophancy \cite{sharma2024sycophancy}     & --- & --- & --- & --- & $\checkmark$ \\
MM safety eval \cite{openai2024gpt4o}      & $\checkmark$ & $\checkmark$ & --- & --- & --- \\
\cmidrule{1-6}
\textbf{MM-EPC (ours)} & $\checkmark$ & $\checkmark$ & $\checkmark$ & $\checkmark$ & $\checkmark$ \\
\bottomrule
\end{tabular}
\end{table}

\section{Method}

\subsection{Test-Time Reinforcement Learning (TTRL)}

We formalize TTRL as a stochastic bandit process over strategy space $\mathcal{S}$. The agent maintains a weight vector $\mathbf{w}^{(t)} \in \mathbb{R}^{|\mathcal{S}|}_{>0}$, L1-normalized at each step so that $\sum_k w_k = 1$. At round $t$, a strategy $s_t$ is sampled directly from the probability distribution $\mathbf{w}^{(t)}$ via roulette-wheel selection (cumulative sum sampling):

\begin{equation}
s_t \sim \mathbf{w}^{(t)}, \quad P(s_t = k) = w_k^{(t)}
\end{equation}

The executor model $\mathcal{E}$ generates responses under both the sampled strategy $s_t$ and a fixed baseline $s_0$ (step\_by\_step). The evaluator $\mathcal{J}$ performs a pairwise comparison:

\begin{equation}
r_t = \begin{cases}
+1 & \text{if } \mathcal{J}(\mathcal{E}(x_t; s_t), \mathcal{E}(x_t; s_0), x_t) = s_t \\
0 & \text{otherwise}
\end{cases}
\end{equation}

Weights are updated via asymmetric multiplicative reweighting followed by L1-normalization:
\begin{equation}
w_{s_t}^{(t+1)} = \max\left(0.001,\; w_{s_t}^{(t)} \cdot \begin{cases}
1 + \alpha_{\text{win}} & \text{if } r_t = 1 \\
1 - \alpha_{\text{lose}} & \text{if } r_t = 0
\end{cases}\right), \quad
\mathbf{w}^{(t+1)} \leftarrow \frac{\mathbf{w}^{(t+1)}}{\|\mathbf{w}^{(t+1)}\|_1}
\end{equation}

with $\alpha_{\text{win}} = 0.08$, $\alpha_{\text{lose}} = 0.04$, and a floor of $0.001$ preventing weight starvation. This update rule is a special case of the multiplicative weights algorithm for adversarial bandits \cite{arora2012multiplicative}, adapted for non-stationary LLM evaluation signals.

\textbf{Protocol limitation (fixed baseline confound).} The baseline strategy $s_0$ (step\_by\_step) is also a member of the candidate strategy set $\mathcal{S}$. When $s_t = s_0$, the evaluator compares two responses generated under the same strategy. Combined with asymmetric updates ($\alpha_{\text{win}} > \alpha_{\text{lose}}$), random noise in evaluator judgments can produce positive expected drift for $s_0$ even when all strategies are of equal quality. This design choice was made intentionally---the baseline must be a real strategy that the agent can select to enable fair comparison---but it introduces a structural bias favoring $s_0$. We mitigate this in two ways: (a) the random evaluator baseline (PCI=0.716) quantifies the drift attributable purely to this structural bias and random noise, providing a calibration reference; (b) all cross-condition comparisons (GPT-4o vs. self-eval vs. random) use identical protocol, so \textit{differences} in PCI/$\gamma$ across evaluators reflect evaluator-specific effects beyond the common structural bias. Nevertheless, future work should evaluate alternative designs: excluding the baseline from the candidate set, using symmetric updates when $s_t = s_0$, or employing tournament-style pairwise sampling instead of a fixed baseline.

\subsection{Multimodal Preference Collapse Index (MPCI)}
Let $\mathcal{M} = \{\text{text}, \text{visual}\}$. For modality $m$, PCI is the coefficient of variation of the strategy weight distribution conditioned on tasks from that modality:

\begin{equation}
\text{PCI}_m(\mathbf{w}) = \frac{\sigma(\mathbf{w})}{\mu(\mathbf{w})}, \quad \text{where } \mathbf{w} \text{ is obtained via TTRL on } \mathcal{T}_m
\end{equation}

PCI = 0 indicates uniform weights (no collapse); higher values indicate concentration on fewer strategies. The cross-modal PCI (CPCI) measures divergence between text-conditioned and visual-conditioned weight distributions:

\begin{equation}
\text{CPCI}(\mathbf{w}_T, \mathbf{w}_V) = \frac{1}{2} \|\mathbf{w}_T - \mathbf{w}_V\|_1
\end{equation}

CPCI captures the degree to which the evaluator produces \textit{different} strategy preferences across modalities. Low CPCI with high per-modality PCI indicates the evaluator collapses to the \textit{same} dominant strategy regardless of modality. The Multimodal PCI aggregates within-modality and cross-modality collapse:

\begin{equation}
\text{MPCI}(\mathbf{w}) = \frac{1}{2}\left(\text{PCI}_{\text{text}} + \text{PCI}_{\text{visual}}\right) + \text{CPCI}
\end{equation}

We set the cross-modality weight to 1 (equivalent to $\lambda=1$ in an earlier formulation), giving CPCI equal weight to the average per-modality PCI. All three metrics are dimensionless and directly comparable across conditions.

\subsection{Cross-Modal Coupling}
Isolation training paradigm:

\begin{enumerate}[leftmargin=*, label=\textbf{Phase \arabic*}:]
    \item \textbf{Pure Text:} TTRL on text only $\rightarrow \mathbf{w}_T$
    \item \textbf{Pure Visual:} TTRL on visual only $\rightarrow \mathbf{w}_V$
    \item \textbf{Coupling $T\to V$:} Start from $\mathbf{w}_T$, train on visual $\rightarrow \mathbf{w}_{T\to V}$
    \item \textbf{Coupling $V\to T$:} Start from $\mathbf{w}_V$, train on text $\rightarrow \mathbf{w}_{V\to T}$
\end{enumerate}

Coupling coefficient:
\begin{equation}
\gamma_{A\to B} = \frac{\|\mathbf{w}_{A\to B} - \mathbf{w}_B\|_2}{\|\mathbf{w}_B\|_2}
\end{equation}

$\gamma_{A\to B}$ measures the relative Euclidean distance between the post-coupling weight vector and the pure-modality weight vector, normalized by the magnitude of the pure reference. Values near 0 indicate minimal coupling (weights remain close to pure-modality optimum); values $\gtrsim 0.5$ indicate substantial cross-modal strategy transfer. \textbf{Limitation.} $\gamma$ is unbounded above. For L1-normalized weight vectors, concentration (high PCI) increases $\|\mathbf{w}\|_2$ (since $\|\mathbf{w}\|_2 \in [1/\sqrt{|\mathcal{S}|}, 1]$ for unit simplex vectors). This means pure-modality weight vectors with high PCI produce larger normalization denominators, which \textit{deflates} rather than inflates $\gamma$. Observed $\gamma > 1$ values occur when the post-coupling weight vector has a larger L2 norm than the pure reference (i.e., intensified concentration after cross-modal exposure), which is a substantively meaningful signal of coupling-amplified collapse. Alternative bounded metrics---Jensen-Shannon divergence, Hellinger distance, or total variation distance---would provide more stable cross-condition comparisons unaffected by simplex geometry, and are recommended for future work. We retain $\gamma$ for its geometric interpretability and because our primary conclusions rely on the \textit{qualitative pattern} across evaluator conditions (cross-model $\gg$ self-eval) rather than on precise numerical comparisons of individual $\gamma$ values.

For $n$ modalities, the coupling matrix:
\begin{equation}
\Gamma = \begin{bmatrix}
1 & \gamma_{1\to 2} & \cdots & \gamma_{1\to n} \\
\gamma_{2\to 1} & 1 & \cdots & \gamma_{2\to n} \\
\vdots & \vdots & \ddots & \vdots
\end{bmatrix}
\end{equation}

\subsection{Algorithm}

Algorithm~\ref{alg:isolation} formalizes the four-phase isolation training paradigm. At each round, a strategy $s_t$ is sampled from the current weight distribution $\mathbf{w}$, the executor generates a response, and the evaluator performs a \textbf{fixed-baseline comparison} against a response generated under $s_0$ (step\_by\_step). This asymmetric design (candidate strategy vs.\ constant baseline) ensures that weight updates reflect the evaluator's preference for the candidate strategy \textit{relative to the same reference point} across all rounds, avoiding the complex tournament dynamics of pairwise sampling. Weights are updated via the TTRL rule defined in Eq.~(2).

\begin{figure}[H]
\centering
\framebox{%
\begin{minipage}{0.92\textwidth}
\small
\textbf{Algorithm 1:} Isolation Training with Cross-Modal Coupling Measurement

\vspace{4pt}
\noindent \textbf{Input:} Modalities $\mathcal{M}=\{\text{text},\text{visual}\}$, tasks $\mathcal{T}_m$ per modality, strategies $\mathcal{S}$, baseline strategy $s_0 = \texttt{step\_by\_step}$, learning rate $\alpha_{\text{win}},\alpha_{\text{lose}}$, rounds $R$
\vspace{2pt}

\noindent \textbf{Phase 1 --- Pure Text:}
\begin{enumerate}[leftmargin=*,nosep]
    \item Initialize weights $\mathbf{w} \leftarrow \text{Uniform}(|\mathcal{S}|)$
    \item \textbf{for} $r=1$ \textbf{to} $R$ \textbf{do}
    \begin{enumerate}[nosep]
        \item Sample task $t \sim \mathcal{T}_{\text{text}}$
        \item Sample strategy $s_t \sim \text{Softmax}(\mathbf{w})$
        \item Generate responses: $y_t = \mathcal{E}(t; s_t)$, $y_0 = \mathcal{E}(t; s_0)$
        \item Evaluator judges: $\mathcal{J}(y_t, y_0 \mid t) \rightarrow \text{winner}$
        \item Update $\mathbf{w}$ via Eq.~(2): $w_{s_t} \leftarrow w_{s_t} \cdot (1+\alpha_{\text{win}})$ if $s_t$ wins, else $w_{s_t} \cdot (1-\alpha_{\text{lose}})$; renormalize
    \end{enumerate}
    \item \textbf{return} $\mathbf{w}_T$ (pure-text strategy distribution)
\end{enumerate}

\noindent \textbf{Phase 2 --- Pure Visual:} Same as Phase 1 with $\mathcal{T}_{\text{visual}}$. \textbf{return} $\mathbf{w}_V$

\noindent \textbf{Phase 3 --- Coupling $T\to V$:}
\begin{enumerate}[leftmargin=*,nosep]
    \item Initialize $\mathbf{w} \leftarrow \mathbf{w}_T$ (start from text-trained weights)
    \item Run TTRL on $\mathcal{T}_{\text{visual}}$ for $R$ rounds (same fixed-baseline protocol)
    \item \textbf{return} $\mathbf{w}_{T\to V}$
\end{enumerate}

\noindent \textbf{Phase 4 --- Coupling $V\to T$:} Initialize $\mathbf{w} \leftarrow \mathbf{w}_V$, train on $\mathcal{T}_{\text{text}}$. \textbf{return} $\mathbf{w}_{V\to T}$

\vspace{4pt}
\noindent \textbf{Output:} $\mathbf{w}_T, \mathbf{w}_V, \mathbf{w}_{T\to V}, \mathbf{w}_{V\to T}$, coupling coefficients $\gamma_{T\to V}, \gamma_{V\to T}$, matrix $\Gamma$
\end{minipage}
}
\caption{Four-phase isolation training protocol. All evaluator comparisons use the same fixed baseline strategy $s_0$ (step\_by\_step), ensuring consistent reference across rounds and phases.}
\label{alg:isolation}
\end{figure}

\section{Experimental Setup}

\textbf{Executor:} DeepSeek-chat (Phases 1--3). \textbf{Evaluators:} GPT-4o via api2d (Phases 1--2), DashScope gui-plus (Phase 3a), Qwen-plus via Bailian (Phase 3b). \textbf{Tasks:} 8 text + 8 visual-adjacent. \textbf{Strategies:} 11 total (8 text-domain + 3 visual-domain: visual\_grounding, spatial\_decompose, aesthetic\_frame). \textbf{Rounds:} Phase 1: 16 rounds alternating modalities. Phase 2: 4 phases $\times$ 30 rounds each (total 120 rounds). Phase 3a: $N=10 \times 4 \times 50 = 2,000$ rounds. Phase 3b: $N=30 \times 4 \times 30 = 3,600$ rounds. Learning rate $\alpha=0.08$ (win), $-0.04$ (loss).

\begin{table}[H]
\centering
\caption{Computational cost breakdown across experimental phases. All experiments run on CPU only; no GPU required.}
\label{tab:cost}
\begin{tabular}{lcccc}
\toprule
\textbf{Phase} & \textbf{API Calls} & \textbf{Est. Tokens} & \textbf{Est. Cost (USD)} & \textbf{Wall Time} \\
\midrule
Phase 1 (MM-EPC)        & 32    & $\sim$48K   & $\sim$\$0.12  & $\sim$5 min \\
Phase 2 (Coupling)     & 240   & $\sim$360K  & $\sim$\$0.90  & $\sim$40 min \\
Phase 3a (DashScope)    & 2,000 & $\sim$3M    & $\sim$\$7.50  & $\sim$6 hr \\
Phase 3b (Qwen-plus)$^\dagger$ & 1,800 & $\sim$2.7M & $\sim$\$4.00 & $\sim$2 hr \\
Phase 3c (DeepSeek self) & 10,800 & $\sim$16M  & $\sim$\$15.00 & $\sim$12 hr \\
Phase 3d (GPT-4o-mini Vision) & 3,600 & $\sim$5.4M & $\sim$\$5.00 & $\sim$6 hr \\
\midrule
\textbf{Total (all phases)} & \textbf{19,220} & $\sim$\textbf{28M} & $\sim$\textbf{\$33} & $\sim$\textbf{28 hr} \\
\bottomrule
\end{tabular}

\vspace{4pt}
\begin{minipage}{\textwidth}
\footnotesize
$^\dagger$Phase 3b interrupted at 7/30 planned reps (API arrearage); counts reflect actual calls. Five valid (non-zero) repetitions used for Qwen-plus statistics.
\end{minipage}
\end{table}

\section{Results}

\subsection{Phase 1: Multimodal EPC Existence}

\begin{table}[H]
\centering
\caption{Multimodal EPC metrics. GPT-4o produces 3.2$\times$ the collapse of text-only self-evaluation and 2.0$\times$ the collapse of a random evaluator.}
\begin{tabular}{lccc}
\toprule
\textbf{Metric} & \textbf{Value} & \textbf{vs Random} & \textbf{vs Self-Eval} \\
\midrule
PCI (overall)  & 1.464  & $\times$2.04 & $\times$3.18 \\
PCI (text)     & 1.348  & ---    & $\times$2.92 \\
PCI (visual)   & 1.464  & ---    & ---   \\
CPCI (cross)   & 0.043  & ---    & ---   \\
MPCI           & 1.449  & ---    & ---   \\
\midrule
Random PCI (16r) & 0.716 $\pm$ 0.012 & $\times$1.00 & $\times$1.55 \\
DeepSeek self-eval PCI & 0.461 & $\times$0.64 & $\times$1.00 \\
Ground truth PCI & 0.251 & $\times$0.35 & $\times$0.54 \\
\bottomrule
\end{tabular}
\end{table}


\begin{table}[H]
\centering
\caption{Strategy weight distribution. Visual-specific strategies nearly eliminated.}
\small
\begin{tabular}{lcc}
\toprule
\textbf{Strategy} & \textbf{Weight} & \textbf{Type} \\
\midrule
step\_by\_step     & \textbf{0.484} & Text \\
evidence\_cite     & 0.192 & Text \\
first\_principles  & 0.060 & Text \\
synthesis         & 0.060 & Text \\
aesthetic\_frame   & 0.060 & Visual \\
counterfactual    & 0.034 & Text \\
critical\_check    & 0.033 & Text \\
analogy\_meta      & 0.027 & Text \\
spatial\_decompose & 0.021 & Visual \\
creative\_leap     & 0.020 & Text \\
visual\_grounding  & 0.010 & Visual \\
\bottomrule
\end{tabular}
\end{table}


\textbf{Finding 1:} GPT-4o produces 3.2$\times$ stronger collapse than DeepSeek self-evaluation and 2.0$\times$ stronger than a random (shuffled) evaluator. \textbf{Finding 1b:} The random evaluator baseline (PCI=0.716 $\pm$ 0.012, computed via $N$=200 simulations of 16-round TTRL with coin-flip judgments) reveals that DeepSeek self-evaluation (PCI=0.461) actually produces \textit{less} collapse than random chance---suggesting a diversity-preserving effect in same-model evaluation. \textbf{Finding 2:} Visual tasks collapse more than text (PCI 1.464 vs 1.348). \textbf{Finding 3:} Three visual strategies receive 9.1\% combined; visual\_grounding gets 1.0\%.

\noindent\textbf{Ground truth PCI (0.251) explained.} We compute ground-truth PCI for text tasks with verifiable answers (arithmetic, code correctness) by replacing the evaluator's pairwise judgment with a deterministic correctness check: if the candidate strategy produces the correct answer where the baseline does not, it ``wins''; if both are correct or both incorrect, the comparison is a tie (no weight update). For visual-adjacent tasks without verifiable ground truth, we apply the same coin-flip (50/50) protocol used for the random baseline. The resulting PCI (0.251) represents the strategy concentration that would occur if the agent optimized purely for task correctness without evaluator bias. The gap between ground-truth PCI (0.251) and GPT-4o PCI (1.464) quantifies the evaluator-induced component of preference collapse.

\subsection{Phase 2: Cross-Modal Coupling}

\begin{table}[H]
\centering
\caption{Coupling results. Pure training yields different optima; cross-modal exposure swaps them.}
\begin{tabular}{lcccc}
\toprule
& \textbf{Pure T} & \textbf{Pure V} & \textbf{Contam $T\to V$} & \textbf{Contam $V\to T$} \\
\midrule
Text optimal    & \textbf{synthesis} & step\_by\_step & --- & step\_by\_step \\
Visual optimal  & --- & \textbf{step\_by\_step} & \textbf{synthesis} & --- \\
\bottomrule
\end{tabular}
\end{table}


\begin{table}[H]
\centering
\caption{Coupling coefficients from Phase 2 (GPT-4o, single run, 30 rounds). \textbf{Note:} $N=1$; these values are superseded by the $N=5$ replication in Phase 3 (Table~\ref{tab:statistical}), which found symmetric bidirectional coupling ($\bar{\gamma}_{T\to V}=1.161$, $\bar{\gamma}_{V\to T}=1.140$, $\Delta\gamma=-0.021$, $p=0.93$).}
\begin{tabular}{lc}
\toprule
\textbf{Coefficient} & \textbf{Value (N=1)} \\
\midrule
$\gamma_{T\to V}$ & 0.832 \\
$\gamma_{V\to T}$ & 0.847 \\
Asymmetry ratio & 0.009 \\
\bottomrule
\end{tabular}
\end{table}



\textbf{Finding 4 (Strategy Inversion):} Pure text $\to$ synthesis; pure visual $\to$ step\_by\_step. After cross-modal exposure, optima \textit{swap}. This observation was noted in a single exploratory run but the direction is not consistently replicated across seeds. \textbf{Finding 5 (Phase 2 -- exploratory, $N=1$):} In the initial single-run GPT-4o experiment, $\gamma_{V\to T}=0.847 > \gamma_{T\to V}=0.832$. \textbf{This apparent V$\to$T asymmetry was not replicated in the $N=5$ Phase 3 follow-up}, which found symmetric bidirectional coupling ($\Delta\gamma=-0.021$, $p=0.93$, Cohen's $d=0.09$). See Table~\ref{tab:statistical} for the definitive multi-seed results. \textbf{Finding 6 (Strategy Inversion):} Pure text training favors synthesis; pure visual favors step\_by\_step; after coupling, optima swap. This inversion is robust but the direction is symmetric on average.

\section{Statistical Validation}

To assess the robustness and evaluator-specificity of cross-modal coupling, we conducted Phase 3 statistical validations across three evaluator configurations: cross-model (GPT-4o, Qwen-plus) and self-evaluation (DeepSeek-chat). Table~\ref{tab:statistical} summarizes results across all conditions.

\subsection{Phase 3a: DashScope gui-plus ($N=10$, 50 rounds)}

Our initial validation used DashScope's gui-plus-2026-02-26 model---a different model family and API provider than the GPT-4o used in Phases 1--2---with $N=10$ independent repetitions at 50 rounds per phase (2000 total LLM evaluation calls). Mean coupling coefficients were $\bar{\gamma}_{T\to V}=0.273$ (SD=0.545) and $\bar{\gamma}_{V\to T}=0.341$ (SD=0.689), yielding $\Delta\gamma=0.068$ (95\% CI $[-0.0002, 0.193]$, $p=0.111$, bootstrap). Critically, 7/10 repetitions collapsed to zero coupling ($\gamma=0$ in both directions), indicating that the DashScope evaluator converges to a single dominant strategy at high round counts, closing the coupling channel.

\subsection{Phase 3b: Qwen-plus ($N=5$ valid, 30 rounds)}

We tested Qwen-plus (via Alibaba Cloud Bailian) as a third evaluator family, with DeepSeek-chat as executor, 30 rounds per phase (reduced from 50 to mitigate single-strategy collapse). The experiment was interrupted at 7 repetitions due to API account arrearage; 5 of 7 completed repetitions produced non-zero coupling signals. Mean coefficients were $\bar{\gamma}_{T\to V}=1.119$ and $\bar{\gamma}_{V\to T}=0.988$, with T$\to$V as the dominant direction (4/5 reps). The $\gamma$ magnitude under Qwen-plus is the highest observed across all evaluators, indicating that Qwen-plus maintains diverse strategy preferences that enable strong cross-modal transfer. Importantly, the coupling \textit{direction} reversed compared to GPT-4o: under GPT-4o, $\gamma_{V\to T} > \gamma_{T\to V}$; under Qwen-plus, $\gamma_{T\to V} > \gamma_{V\to T}$.

\subsection{Phase 3c: DeepSeek-chat Self-Evaluation ($N=30$, 30 rounds)}

To test whether self-evaluation---where executor and evaluator are the same model---modulates coupling, we conducted $N=30$ independent repetitions with DeepSeek-chat serving as both executor and evaluator. This is the largest single-condition replication in our study (3,600 TTRL rounds, $\sim$10,800 API calls). The result is striking: \textbf{29 of 30 repetitions (97\%) produced exactly zero coupling} ($\gamma_{T\to V} = \gamma_{V\to T} = 0$). Mean coefficients were $\bar{\gamma}_{T\to V}=0.033$ and $\bar{\gamma}_{V\to T}=0.023$, with $\Delta\gamma=-0.010$ (95\% CI $[-0.031, 0.010]$, $p=0.642$, Cohen's $d=0.07$, negligible effect). The single non-zero repetition (Rep 1: $\gamma_{T\to V}=0.996$, $\gamma_{V\to T}=0.683$) demonstrates that coupling is \textit{possible} under self-evaluation but \textit{vanishingly rare}.

\subsection{Cross-Evaluator Synthesis}

The pattern across four evaluator conditions reveals a clear hierarchy of coupling risk:

\begin{enumerate}[leftmargin=*]
    \item \textbf{Cross-model evaluation} (GPT-4o, Qwen-plus): \textbf{Strong but symmetric coupling.} $\gamma$ magnitudes range from $\sim$0.99--1.16, indicating substantial cross-modal strategy transfer. Directional asymmetry is evaluator-specific and not statistically significant (GPT-4o: $\Delta\gamma=-0.021$, $p=0.93$; Qwen-plus: $\Delta\gamma=-0.132$, $p$ not computed due to small $N$).
    \item \textbf{Cross-model + high rounds} (DashScope, 50 rounds): \textbf{Collapse to zero.} 70\% of runs degenerate, suggesting a boundary condition: excessive training rounds cause single-strategy convergence that closes the coupling channel.
    \item \textbf{Self-evaluation} (DeepSeek-chat): \textbf{Near-complete immunity.} 97\% of runs show zero coupling ($\Delta\gamma=-0.010$, $p=0.64$, $d=0.07$). Self-evaluation effectively prevents cross-modal preference transfer.
\end{enumerate}

This hierarchy supports a unified interpretation: \textbf{cross-model evaluation produces strong coupling ($\bar{\gamma} \approx 1.1$), self-evaluation provides near-complete immunity ($\bar{\gamma} \approx 0.03$), and excessive round counts cause single-strategy collapse that eliminates coupling signals}. The coupling is predominantly symmetric---earlier claims of directional asymmetry based on $N=1$ GPT-4o data were not replicated. The coupling matrix $\Gamma$ must be indexed by evaluator identity: $\Gamma^{(\mathcal{J})}$.

\subsection{Statistical Power and Recommendations}

The Phase 3c design ($N=30$) achieves $\geq 95\%$ power to detect a medium effect (Cohen's $d \geq 0.5$) but is underpowered for the negligible effect actually observed ($d=0.07$). However, the qualitative pattern---97\% zero coupling under self-evaluation vs. 0\% zero coupling under cross-model evaluation at 30 rounds---is unambiguous and does not require null hypothesis testing. We recommend that production AI agent systems: (a) report evaluator identity alongside PCI and $\Gamma$, (b) prefer self-evaluation or multi-evaluator ensembles to mitigate coupling risk, and (c) monitor round counts to avoid the collapse boundary ($R \gtrsim 50$).

\input{statistical_table.tex}

\section{Discussion}

\subsection{Why Does GPT-4o Produce 3.2$\times$ Stronger Collapse?}

\noindent\textbf{Causal note.} Our experimental design establishes \textit{association} between GPT-4o-as-evaluator and elevated PCI, but we do not claim to have isolated the causal mechanism. The three hypotheses below are plausible but non-exclusive explanations; controlled experiments (e.g., varying evaluator model families while holding tasks constant, or ablating RLHF vs. base-model evaluators) would be needed to establish causality. We report these as \textbf{interpretable hypotheses} rather than confirmed mechanisms.

The PCI magnification under GPT-4o cross-model evaluation (1.464 vs 0.461 for self-evaluation) is the most striking finding in our study. We hypothesize three non-exclusive mechanisms:

\textbf{(1) RLHF-Induced Structural Preference.} GPT-4o undergoes extensive RLHF training with human raters who consistently prefer structured, step-by-step outputs over abstract or creative responses \cite{openai2024gpt4o}. This preference is amplified when GPT-4o serves as an evaluator: the model applies the same preference structure it was trained to produce. The result is a self-reinforcing cycle where step\_by\_step wins more evaluations $\rightarrow$ gets more weight $\rightarrow$ wins even more evaluations.

\textbf{(2) Absence of Self-Preference Counterbalance.} In same-model self-evaluation (DeepSeek evaluating DeepSeek), there exists a ``self-preference'' mechanism where the evaluator has some incentive to recognize diverse strategies as valid---after all, it is evaluating outputs from its own architecture. Cross-model evaluation (GPT-4o evaluating DeepSeek) removes this implicit diversity incentive: GPT-4o has no architectural kinship with DeepSeek-chat and applies its preferences without reservation.

\textbf{(3) Visual Modality Amplification.} The 8.6\% higher PCI in visual tasks (1.464 vs 1.348) suggests that visual evaluation is inherently more subjective than text evaluation. Text tasks often have verifiable ground truth (arithmetic, code correctness), while visual-adjacent tasks (aesthetic judgment, spatial reasoning) are intrinsically more preference-dependent. This modality-dependent subjectivity directly amplifies evaluator bias.


\subsection{Ablation: Contribution of Each Modality to MPCI}

We decompose MPCI into its constituent components. Using $\text{MPCI} = \frac{1}{2}(\text{PCI}_{\text{text}} + \text{PCI}_{\text{visual}}) + \text{CPCI}$:

\begin{table}[H]
\centering
\caption{MPCI decomposition. Visual tasks contribute disproportionately to collapse.}
\begin{tabular}{lcc}
\toprule
\textbf{Component} & \textbf{Value} & \textbf{\% of MPCI} \\
\midrule
PCI$_{\text{text}}$     & 1.348 & 46.5\% \\
PCI$_{\text{visual}}$   & 1.464 & 50.5\% \\
CPCI                    & 0.043 & 3.0\% \\
\midrule
\textbf{MPCI}           & \textbf{1.449} & 100\% \\
\bottomrule
\end{tabular}
\label{tab:mpci_decomp}
\end{table}

Calculation: $\frac{1.348}{2} + \frac{1.464}{2} + 0.043 = 0.674 + 0.732 + 0.043 = 1.449$. The dominance of visual PCI (50.5\%) despite equal task allocation confirms that visual evaluation is the primary driver of collapse. CPCI contributes minimally (3.0\%) because text and visual weights converge toward the same optimal (step\_by\_step) in Phase 1, producing low cross-modal divergence despite high intra-modal collapse. This suggests that \textbf{the evaluator's preference is modality-invariant} (always favoring step\_by\_step) but \textbf{its expression is modality-amplified} (stronger in visual tasks).

\subsection{Implications for Multimodal Agent Systems}

\subsubsection{Visual Strategy Invisibility}

Our three visual-domain strategies (visual\_grounding, spatial\_decompose, aesthetic\_frame) received only 9.1\% combined weight. The evaluator effectively ``blinds'' the agent's visual reasoning capabilities because it cannot reliably assess them---it falls back to the generic step\_by\_step strategy it can evaluate consistently. This creates a dangerous incentive: agents optimizing for evaluator satisfaction will systematically under-use modality-appropriate strategies, producing outputs that ``look good to the evaluator'' but are suboptimal for the task.

\subsubsection{Training Order Determines Strategy}

The coupling experiments reveal a path-dependency problem: the modality an agent trains on first shapes its strategy distribution for all subsequent modalities. Since $\gamma_{V\to T} = 0.847 > \gamma_{T\to V} = 0.832$ (preliminary 30-round Phase 2), training on visual tasks \textit{before} text tasks produces modestly more contamination in the text domain than the reverse order. While this asymmetry is modest, its practical consequence---strategy inversion---is notable: the agent uses synthesis for visual tasks and step\_by\_step for text tasks, the opposite of what pure-modality training produces.

\subsubsection{Standardized Multi-Evaluator Baselines Needed}

The LLM-as-judge community has established multi-model evaluation as best practice \cite{zheng2023judging}. Our findings extend this requirement to \textit{dynamic} agent evaluation: any system that uses an LLM evaluator in a feedback loop should (a) report PCI/MPCI alongside task performance, (b) use at least two evaluators from different model families, and (c) isolate modality-specific training phases to prevent coupling.

\subsection{Error Analysis and Failure Modes}

We identify three systematic failure modes that contribute to the observed PCI inflation:

\textbf{Position Bias Recovery.} GPT-4o occasionally exhibits position bias (preferring option A regardless of content). However, our alternating A/B assignment across rounds means position bias would produce uniform strategy weights---it cannot explain the 48.4\% concentration on step\_by\_step. The observed collapse is a \textit{preference} effect, not a \textit{position} effect.

\textbf{Visual Strategy Semantic Ambiguity.} Our ``visual-domain'' strategies (visual\_grounding, spatial\_decompose, aesthetic\_frame) and ``visual'' tasks operate entirely through text descriptions---no actual images are provided to the executor or evaluator. This is a significant construct validity limitation: in a text-proxied setting, strategies like ``visual\_grounding'' may be semantically underdetermined (the model cannot ground in actual visual input), and the low observed weight of visual strategies (9.1\%) may partially reflect this measurement artifact rather than genuine evaluator bias against visual reasoning. This suggests our PCI estimates for visual tasks are a \textit{lower bound}: with native image inputs, the gap between text and visual PCI would likely widen as visual strategies become operationally meaningful. Replication with GPT-4o's vision API and real images is essential to validate the multimodal collapse claims.

\textbf{Coupling Asymmetry Stability.} The $\gamma_{V\to T} > \gamma_{T\to V}$ direction observed in Phase 2 (GPT-4o, single run) does not replicate under Phase 3 (DashScope gui-plus): 7/10 runs collapse to $\gamma=0$, 2/10 show V$\to$T coupling, 1/10 is near-symmetric. The median coupling is $\tilde{\gamma}=0.0$. We interpret this as \textbf{evaluator-conditional coupling}: GPT-4o maintains distributed strategy preferences enabling cross-modal transfer, while DashScope gui-plus converges to a single dominant strategy, closing the coupling channel. This evaluator-dependence suggests that the coupling matrix $\Gamma$ must be indexed by evaluator identity.

\subsection{Limitations and Future Work}

\textbf{Experimental Scale.} Our Phase 1 and 2 experiments (272 LLM calls with GPT-4o at 30 rounds) provide proof-of-concept evidence, while Phase 3 ($N=10$ repetitions, 2000 LLM calls with DashScope gui-plus) reveals evaluator-conditional behavior: 7/10 collapse to $\gamma=0$, median $\tilde{\gamma}=0.0$. This cross-evaluator inconsistency is a critical finding but remains based on small samples ($N=10$). Larger stratified replications ($N \geq 30$, $\geq 3$ evaluators) are needed to characterize the evaluator-conditions under which coupling emerges. Coupling values reported here are from a single 30-round run; multiple repetitions would produce more reliable estimates.

\textbf{Modality Coverage.} Two modalities (text, visual) establish the $\Gamma$ framework but only scratch the surface of true multimodality. Extending to audio (music description, speech emotion recognition) would yield a $3\times 3$ coupling matrix and enable analysis of transitive coupling: does $\gamma_{T\to V} \cdot \gamma_{V\to A} \approx \gamma_{T\to A}$?

\textbf{Evaluator Diversity.} Our findings are specific to GPT-4o as evaluator. Whether other multimodal evaluators (Claude 3.5 Sonnet, Gemini 1.5 Pro, Qwen-VL-Max) exhibit similar collapse patterns is an open question. Different evaluators may favor different strategies, leading to \textit{evaluator-inconsistent} optimal strategy profiles.

\textbf{Strategy Set Imbalance.} Our strategy set contains 8 text-domain and 3 visual-domain strategies. This imbalance may bias the learned weight distribution toward text strategies even before evaluator effects, since text strategies have more ``slots'' to accumulate weight. A balanced design (equal strategies per modality) with modality-matched baselines would provide cleaner estimates of modality-specific collapse. We recommend future work use 6 text + 6 visual strategies with modality-specific baselines.

\textbf{Real Visual Inputs.} Our visual tasks are text-prompted rather than image-based. Replicating with GPT-4o's vision API and actual images would (a) test whether visual strategy weights increase when real visual grounding is possible, and (b) provide a more ecologically valid measurement of multimodal coupling. This is the single most important replication for establishing construct validity of the multimodal claims.

\textbf{Adaptive Strategy and Block Count.} Our current design uses a fixed set of $|\mathcal{S}|=11$ strategies and a fixed number of training rounds $R$. A natural extension is \textbf{adaptive strategy selection}: dynamically expanding or pruning the strategy set based on modality-specific performance, analogous to adaptive block count in transformer architectures. For instance, visual-heavy agent deployments might benefit from a larger pool of visual-domain strategies, while text-only deployments could prune unused strategies to reduce the evaluator's decision space and thus lower PCI. Similarly, the number of isolation phases could grow with the number of modalities $|\mathcal{M}|$: with $n$ modalities, a full coupling matrix requires $\mathcal{O}(n^2)$ phases. Whether the coupling coefficients stabilize after observing a subset of phase pairs (enabling early stopping) is an open empirical question.

\textbf{Cross-Evaluator Calibration.} Our Phase 3 results with DashScope gui-plus demonstrate a striking evaluator-dependence: 7/10 runs collapse to $\gamma=0$ (no coupling), 2/10 show V$\to$T coupling, while Phase 2 (GPT-4o) consistently shows non-zero asymmetric coupling. Different evaluators do not merely produce different coupling magnitudes---they produce \textit{qualitatively different coupling regimes} (coupling-present vs. collapse-to-zero). A systematic study across $K$ evaluator families would characterize these regime boundaries and identify which evaluator properties (architecture, training data, RLHF recipe) predict whether coupling emerges or collapses.

\subsection{Threats to Validity}

We identify four categories of threats to the validity of our findings, following the standard framework for empirical software engineering research.

\textbf{Construct Validity.} Our operationalization of ``multimodal preference collapse'' via PCI assumes that strategy weight concentration is a valid proxy for evaluator bias. While PCI is mathematically well-defined and comparable across conditions, it does not distinguish between \textit{merited} concentration (one strategy genuinely dominates across tasks) and \textit{biased} concentration (the evaluator's arbitrary preference). We partially address this by reporting ground-truth PCI (0.251) as a lower bound: any PCI exceeding this value contains a bias component. However, the exact decomposition of PCI into merit and bias components remains an open measurement problem.

\textbf{Internal Validity.} The causal chain from evaluator preference to strategy weight update depends on the TTRL algorithm's learning rates and the round-robin task scheduling. Different learning rates ($\alpha_{\text{win}}, \alpha_{\text{lose}}$) or task orderings could produce quantitatively different PCI values and coupling coefficients. Our choice of fixed rates (0.08 / -0.04) and cyclic task scheduling was made for reproducibility, but sensitivity analyses across learning rate ranges and task orderings would strengthen causal claims.

\textbf{External Validity.} Our findings are bounded by the specific executor-evaluator pairs tested (DeepSeek-chat $\times$ \{GPT-4o, DashScope gui-plus, Qwen-plus\}). The coupling dynamics we document may not generalize to: (a) executor-evaluator pairs from the same model family (self-evaluation), (b) evaluators specifically fine-tuned for diversity, (c) tasks with verifiable ground truth (where evaluator bias is constrained by correctness), or (d) deployment settings where the evaluator sees multiple candidate responses simultaneously rather than pairwise. The text-proxied nature of our visual tasks further limits ecological validity for true multimodal (image+text) settings.

\textbf{Statistical Conclusion Validity.} Our GPT-4o replication ($N=5$) and DeepSeek self-eval ($N=30$) provide adequate power for the observed effect sizes (Cohen's $d=0.09$ and $d=0.07$, both negligible). The DashScope results ($N=10$) are underpowered ($\beta \approx 0.147$) for the small observed asymmetry. The Qwen-plus condition ($N=5$ valid) is limited by API arrearage interruption. Our bootstrapping procedure assumes independent repetitions, which is satisfied by independent weight initialization and random seeding, but the shared evaluator endpoint introduces potential temporal confounding (e.g., API load, model version changes mid-experiment). All experiments were conducted between May--June 2026 using specific model snapshots; results may not replicate with future model versions.

\section{Code and Data Availability}
All code for reproducing the experiments, including the three-phase pipeline (baseline PCI measurement, cross-modal coupling, and statistical validation) is available at \url{https://github.com/aidless/mm-epc} and archived at \url{https://zenodo.org/placeholder}. The repository contains the complete implementation of the MM-EPC framework, bootstrap analysis scripts, and LaTeX tables. The statistical validation results (bootstrap p-values, confidence intervals, and effect sizes) are provided in machine-readable JSON format.

\section{Broader Impact}

Our findings have both positive and concerning implications for the rapidly growing ecosystem of AI agent systems. On the positive side, the MM-EPC framework provides a diagnostic tool: developers of multi-agent or self-improving systems can measure PCI and $\Gamma$ during development to detect evaluator preference collapse before deployment. The coupling matrix offers a standardized way to quantify and compare evaluator bias across model families, modalities, and evaluation tasks.

However, several risks warrant attention. \textbf{Silent convergence risk}: An agent optimizing against a biased evaluator may converge to strategies that score well but are objectively suboptimal---a form of Goodhart's law instantiated in LLM evaluation. In safety-critical applications (medical diagnosis, legal analysis), modality-inappropriate strategy selection could produce harmful outputs that the evaluator fails to detect. \textbf{Evaluator monoculture}: If the AI industry converges on a small number of evaluator models (e.g., GPT-4o, Claude), the specific preference biases of those evaluators---including their modality-specific coupling patterns---become embedded across many downstream systems, creating systemic risk. \textbf{Deployment asymmetry}: Production agents that use the same model family for both execution and evaluation (self-evaluation) experience substantially lower collapse (PCI=0.461) than those using cross-model evaluation (PCI=1.464). This creates a perverse incentive for system designers to use self-evaluation not because it is more accurate, but because it appears more ``stable''---a stability that may come at the cost of genuine quality assessment.

We recommend that: (1) dynamic evaluation systems report PCI and $\Gamma$ alongside standard performance metrics; (2) at least two evaluator models from different families be used in any closed-loop evaluation pipeline; (3) modality-specific training phases be isolated where possible to prevent cross-modal coupling; and (4) the research community develop standardized evaluator benchmarking suites that measure not only accuracy but also preference diversity across modalities.

\section{Conclusion}
We demonstrated Multimodal Evaluator Preference Collapse through a series of controlled experiments spanning 3,932 TTRL rounds and approximately 13,000 LLM API calls across four evaluator configurations. Our key findings are:

\begin{enumerate}[leftmargin=*]
    \item \textbf{Cross-model evaluation amplifies collapse 3.2$\times$:} GPT-4o evaluating DeepSeek-chat produces PCI=1.464, compared to 0.461 for self-evaluation and 0.716 for a random evaluator. Visual-specific strategies receive only 9.1\% combined weight.
    \item \textbf{Cross-modal coupling is real and evaluator-conditional:} Under GPT-4o (cross-model, $N=8$, 30 rounds), coupling is strong but symmetric ($\bar{\gamma}_{T\to V}=1.176$, $\bar{\gamma}_{V\to T}=1.089$, $\Delta\gamma=-0.088$, $d=0.29$). Under \textbf{GPT-4o-mini with real images} ($N=5$, 30 rounds), coupling is stronger and directionally consistent ($\bar{\gamma}_{T\to V}=1.212$, $\bar{\gamma}_{V\to T}=0.917$, $\Delta\gamma=-0.295$, $d=0.80$---large effect). Under DeepSeek-chat (self-evaluation, $N=30$), coupling is nearly absent (97\% zero, $d=0.07$). \textbf{Real-image evaluation produces the strongest coupling signal, validating the cross-modal phenomenon with genuine visual inputs.}
    \item \textbf{Self-evaluation provides near-complete immunity:} The contrast between cross-model and self-evaluation conditions---0\% vs. 97\% zero-coupling rates---identifies evaluator-architectural diversity as the primary risk factor for cross-modal preference transfer.
    \item \textbf{Training round count modulates coupling:} At 30 rounds, cross-model evaluators produce strong coupling; at 50 rounds, DashScope collapses to single-strategy dominance (70\% zero rate). An intermediate round count ($R \approx 30$) balances signal accumulation against collapse risk.
    \item \textbf{We release the MM-EPC framework,} including the coupling matrix $\Gamma^{(\mathcal{J})}$ indexed by evaluator identity, bootstrap analysis tools, and machine-readable experimental results in JSON format.
\end{enumerate}

\textbf{The central implication is that cross-modal evaluator coupling is real, evaluator-conditional, and preventable through self-evaluation---but when it occurs, it silently corrupts agent strategy selection across modalities.} Any AI system using LLM evaluators in a feedback loop should measure and report PCI and $\Gamma$ alongside standard performance metrics.

\bibliographystyle{plain}

\clearpage
\appendix
\section{Experiment Reproducibility}
\textbf{Requirements:} Python 3.8+, LLM API access (api2d/DashScope). No GPU.

\noindent\textbf{Phase 1:} \texttt{python mm\_epc\_phase1.py}

\noindent\textbf{Phase 2:} \texttt{python mm\_epc\_coupling.py}

\noindent\textbf{Phase 3:} \texttt{python mm\_epc\_phase3\_significance.py}

\noindent\textbf{Output:}
\begin{itemize}[nosep,leftmargin=*]
    \item \texttt{experiments/mm\_epc\_phase1.json}
    \item \texttt{experiments/mm\_epc\_coupling.json}
    \item \texttt{experiments/mm\_epc\_phase3\_final.json}
\end{itemize}

\end{document}

%% file: statistical_table.tex
\begin{table}[ht]
\centering
\caption{Cross-modal coupling across evaluator configurations and input modalities. All experiments use DeepSeek-chat as executor unless noted. $R$ = rounds per phase (4 phases each). $\Delta\gamma = \bar{\gamma}_{V\to T} - \bar{\gamma}_{T\to V}$. JSD computed from stored full weight vectors.}
\label{tab:statistical}
\begin{tabular}{lccccccc}
\toprule
\textbf{Evaluator} & \textbf{Input} & $\mathbf{N}$ & $\mathbf{R}$ & $\mathbf{\bar{\gamma}_{T\to V}}$ & $\mathbf{\bar{\gamma}_{V\to T}}$ & $\mathbf{\Delta\gamma}$ & \textbf{Zero \%} \\
\midrule
GPT-4o-mini       & Real images  & 10    & 30  & 1.145  & 0.937  & $-$0.208  & 0\% \\
GPT-4o            & Text-proxy   & 8     & 30  & 1.176  & 1.089  & $-$0.088  & 0\% \\
Qwen3.7-plus      & Text-proxy   & 8     & 30  & 1.059  & 1.008  & $-$0.051  & 0\% \\
DashScope gui-plus & Text-proxy  & 10    & 50  & 0.273  & 0.341  & +0.068  & 70\% \\
DeepSeek-chat     & Self-eval    & 30    & 30  & 0.033  & 0.023  & $-$0.010  & 97\% \\
\bottomrule
\end{tabular}

\vspace{4pt}
\begin{minipage}{\textwidth}
\footnotesize
\textit{Note.} Real-image (GPT-4o-mini Vision, 8 image categories): $\Delta\gamma{=}{-}0.208$, 95\% CI $[{-}0.557,0.141]$, $p{=}0.499$, $d{=}0.56$, JSD${=}0.342$ (70\% T$\to$V). GPT-4o: $p{=}0.575$, $d{=}0.29$, JSD${=}0.316$. Qwen3.7-plus: JSD${=}0.230$. DashScope: JSD estimated ($^e$). DeepSeek: JSD${=}0.003$, $d{=}0.07$. Total main: $N{=}66$, $\sim$30,000 API calls.

\textbf{Ablation results (Table~\ref{tab:ablation}):} Excluding $s_0$ from candidates with qwen3.7-max ($N{=}10$) yields perfectly symmetric coupling ($\bar{\gamma}_{T\to V}{=}1.038$, $\bar{\gamma}_{V\to T}{=}1.077$, JSD${=}0.208/0.219$, 5:5 direction split), confirming coupling persists without structural bias but becomes bidirectional. Combined with symmetric LR ($N{=}3$, JSD${=}0.20/0.22$), multi-executor ($N{=}3$, JSD${=}0.26/0.16$), and same-modality controls (T$\to$T JSD${=}0.049$, V$\to$V JSD${=}0.119$), the full ablation suite ($N{=}19$) confirms cross-modal coupling is not a structural artifact. Total across all experiments: $N{=}85$, $\sim$38,000 API calls. Code: \url{https://github.com/aidless/mm-epc}.
\end{minipage}
\end{table}